\documentclass{article}

\usepackage{PRIMEarxiv}

\usepackage[utf8]{inputenc} % allow utf-8 input
\usepackage[T1]{fontenc}    % use 8-bit T1 fonts
\usepackage{hyperref}       % hyperlinks
\usepackage{url}            % simple URL typesetting
\usepackage{booktabs}       % professional-quality tables
\usepackage{amsfonts}       % blackboard math symbols
\usepackage{nicefrac}       % compact symbols for 1/2, etc.
\usepackage{microtype}      % microtypography
\usepackage{lipsum}
\usepackage{fancyhdr}       % header
\usepackage{graphicx}       % graphics
\graphicspath{{media/}}     % organize your images and other figures under media/ folder
\usepackage{array}
\usepackage{algorithm}
\usepackage{algpseudocode}
\usepackage{orcidlink}
\usepackage{caption}
\usepackage{subcaption}

\title{Detecting Evidence of Organization in groups by Trajectories
}

\author{
  \orcidlink{0000-0003-2535-3024}Thayanne França da Silva, \orcidlink{0000-0002-5932-9818}Jos\'e Everardo Bessa Maia \\
  Universidade Estadual do Cear\'a - UECE, 60714-903, Fortaleza-CE, Brasil \\
  \texttt{thayanne.silva@aluno.uece.br, jose.maia@uece.br} \\
}

\begin{document}
\maketitle

\begin{abstract}
Effective detection of organizations is essential for fighting crime and maintaining public safety, especially considering the limited human resources and tools to deal with each group that exhibits co-movement patterns. This paper focuses on solving the Network Structure Inference (NSI) challenge. Thus, we introduce two new approaches to detect network structure inferences based on agent trajectories. The first approach is based on the evaluation of graph entropy, while the second considers the quality of clustering indices. To evaluate the effectiveness of the new approaches, we conducted experiments using four scenario simulations based on the animal kingdom, available on the NetLogo platform: Ants, Wolf Sheep Predation, Flocking, and Ant Adaptation. Furthermore, we compare the results obtained with those of an approach previously proposed in the literature, applying all methods to simulations of the NetLogo platform. The results demonstrate that our new detection approaches can more clearly identify the inferences of organizations or networks in the simulated scenarios.
\end{abstract}

% keywords can be removed

\section{Introduction}
A wide range of research has been devoted to detecting anomalous behavior, whether through video analysis or tracking the trajectories of objects \cite{Coşar}. However, the complexity increases considerably when extending this challenge to detecting group anomalous behavior. Identifying abnormal behavior patterns in public spaces, especially when manifested by groups, presents additional challenges. This advanced detection capability has the potential for enhanced monitoring of public safety levels \cite{Paul}.

Scenes obtained from surveillance video are generally of low resolution, have occlusion occurring, and have a limited field of observation \cite{Paul}. The last few decades have seen steady growth in location-tracking devices (e.g., vehicle navigation systems and smartphones). This has generated a massive amount of trajectory data for co-motion pattern mining \cite{LiKe2023Rgpd}. A co-motion pattern indicates a set of moving objects traveling together over some time \cite{LiKe2023Rgpd}. Some examples of representation of this pattern include flocks \cite{Vieira}, convoys \cite{convoy}, swarms \cite{LiZhenhui2010Smrt}, groups \cite{Yida}, platoons \cite{LI2015167}, and meetings \cite{Zheng}. 

In analyzing movement and behavior patterns, networks are a natural choice of representation, as they can highlight connections between agents, locations, events, or any entity of interest. Networks, composed of nodes and edges that model the relationships between those nodes, offer an effective way to capture the underlying structure and interconnections in data \cite{related1}.

Inferred networks, in particular, form a class whose definitions of nodes and/or edges are derived from the data without the need for predefined relationships or a priori knowledge about the network structure \cite{survey}. This allows previously unnoticed or hidden information to emerge from data analysis, providing new perspectives and insights. This approach is particularly valuable in situations where the interactions between elements are not well understood or when the data is heterogeneous and cannot be easily translated into a traditional network format.

In the context of surveillance in public spaces, where the accurate identification of individuals often takes place through tracking devices to the detriment of traditional security cameras, the trajectory-based Network Structure Inference approach emerges as a promising strategy for the surveillance of public spaces detection of anomalous co-movement patterns in groups.

Consequently, the present study introduces two strategies to identify evidence of organizational(network) structure. To evaluate the effectiveness of these approaches, we conducted tests on four simulations of organizational dynamics in the animal kingdom, which were made available through the NetLogo platform \cite{Uri}. These simulations encompass several combinations of values for the number of agents and scenario dimensions, providing a comprehensive assessment of the performance of our approaches.

The main contributions of this work are summarized as follows:
\begin{itemize}
    \item Two new approach is proposed using the trajectory. One generate clusters of objects with co-motion, then the cluster quality index is evaluated to identify simulations that have evidence of organization (high-quality index). The other using graph entropy calculation to detect behavior anomalies;
    \item To evaluate the proposed approach, we used simulations from the animal kingdom, which are readily accessible in the NetLogo library. These simulations cover scenarios with agents exhibiting organized and disorganized patterns, providing a solid basis for evaluating our methodology. In many contexts, the absence of robust datasets for evaluation is a recurrent issue. However, multi-agent simulations are crucial in filling this gap, allowing a reliable and comprehensive evaluation of proposed approaches.
\end{itemize}

This paper is organized into four more sections. In Section II, presents the related work. Furthermore, it shows the experiment and the results. Section III describes essential concepts in this work, such as the DBSCAN clustering algorithm, Silhouette Coefficient, Network Structure Inference, and Entropy. Section IV, the scenarios, the organization, and our approach are described. Section V shows our results compared to the literature. Finally, Section VI concludes the paper with final observations and future research.

\section{Related Work}

\cite{silva2020} and \cite{silva2021} introduced the problem of classifying types of organizations in multi-agent systems in a scenario where a group of target mobile agents is continuously monitored by a smaller group of mobile observer agents in the CTO problem \cite{maia2021cooperative}. The approach by \cite{silva2020} considers that the group of target agents can be organized according to eight paradigms (hierarchy, holarchy, team, coalition, congregation, society, federation and matrix organizations), while the approach by \cite{silva2021} considers four paradigms (hierarchy, holarchy, teams and coalition).

The approach proposed by \cite{silva2020} consists of collecting the exchange of messages from all agents of the simulation that is shared with the two rival groups, and through seven supervised learning classifiers, the classification of the detected organizational paradigms is carried out.The approach proposed by \cite{silva2021} collects images of the simulation scenario at each time step and through Mobilenetv2 the classification of organizational paradigms is carried out.

The results of both approaches showed that the approaches had satisfactory results in the classification of each organizational paradigm. However, the two scenarios used in the approaches are unrealistic: exchanging messages between agents from different rival groups and identifying the type of organization using only the scenario images.

\cite{related1} presented a generic approach to an entropy-based analysis, which uses the combined and automated analysis of short-term and long-term behavior of entropy values over some time to characterize and examine the self-organizing behavior of systems complexes.

The approach consisted of obtaining the x and y coordinates of each evaluated simulation. Using the values of these discretized parameters, a histogram for each parameter is created \cite{related1}. From probability $pi$, which is the probability of a discretized value of the calculated histogram, entropy is calculated. Finally, for the automated detection of self-organization, the entropy values were calculated using filters, and the time derivative of the entropy values was post-processed as input for the analysis of the short-term and long-term behavior of the systems.
 
\cite{related1} used a small window size (25 simulation time steps) and a larger window size ($10\%$ of all simulation steps) for the analysis of long-term behavior in the evaluation. In addition, two simulation models were used. The first is the chicken simulation model, where a negative self-organized behavior in this scenario occurs if a chicken is injured and the other chickens hunt and surround the injured chicken and try to peck it to death. The second is the pollination model, which simulates the movements of bees as they collect and spread pollen from flowers.

The results of the evaluated systems indicate that the approach proposed by \cite{related1} is adequate since the combined use of short-term and long-term behavior analysis works in identification. However, this approach has not been evaluated in scenarios with a larger number of agents, larger simulation environments, unlimited environments, other forms of organized systems, etc.

\section{Background}

\subsection{DBSCAN Clustering}

O DBSCAN was proposed in 1996 as the first Density-Based Clustering Algorithm (DBCLA) \cite{clustering}. The key idea is that for each cluster point, the neighborhood of a given radius has to contain at least a minimum number of points, i.e., the density in the neighborhood has to exceed some threshold \cite{clustering}. 

Therefore, DBSCAN takes two parameters, $Eps$ and $Minpts$. The $Minpts$ defines the minimum amount of points to define whether a point $p$ is the center in a cluster, an edge, or noise in dataset $D$. While for a given point $p$, $Eps$ signifies the radius of its surrounding region known. The literature denotes $Eps$ neighborhood of $p$ as $N_{Eps}(p)$. Let $D$ denote the dataset, then for any $p \in D$, its $Eps$ neighborhood is given the Equation (1): 

\begin{equation}
N_{Eps}(p) = \{q \in D \| dist(p,q) \leq Eps\}
\end{equation} 

If the $N_{Eps}(p)$ of an object $p$ contains at least a minimum number, $Minpts$, of objects, then the object $p$ is called a core point. If the $N_{Eps}(p)$ has less than $Minpts$, but some core point, object $p$ is called an border point. Otherwise, the point will be considered noise. Two principal points have the same cluster membership if they are directly reachable by density with respect to $Eps$ and $Minpts$ in a set $D$, if there exists a chain of objects ${p_{1},..., p_{n}}$, such that $p_{1}= q$ and $p_{n}= p$ and $p_{i}+1$ is reachable by density directly from pi with respect to e $Minpts$, for $1 \leq i \leq n$, $p_{i}$ in $D$ \cite{clustering}. Algorithm 1 presents the pseudocode of the DBSCAN algorithm.

\begin{algorithm}[H]
  \caption{DBSCAN Algorithm}
  \scriptsize
  \label{algorithmdbscan}
  \begin{algorithmic}[1]
  \Require $X, Eps, MinPts$
    \State Mark all points $x \in X$ as noise
    \For{$unvisited\ point\ x\ \in X$}
        \State Mark $x$ as visited
        \State $N \gets GETNEIGHBORS(x, Eps)$
        \If{$|N| \geq MinPts$}
        \State Mark $x$ as core
            \For{$unvisited\ point\ y\ \in N$}
            \State $M \gets GETNEIGHBORS(y, Eps)$
            \If{$|M| \geq MinPts$}
            \State Mark $y$ as core
            \Else
            \State Mark $y$ as border
            \EndIf
            \EndFor
        \EndIf
    \EndFor
        
  \end{algorithmic}
\end{algorithm}

As each simulation has different numbers, scenarios, and objectives, different numbers of clusters can be formed, we selected the DBSCAN clustering algorithm. In addition, DBSCAN allows focusing on the agents involved in the groups, avoiding noise.

\subsection{Silhouette Coefficient}

There are many algorithms for partitioning a set of objects into clusters, but these clustering methods always result in $k$ clusters, whatever the data. However, it is necessary to assess whether the partitioning reflects a grouping structure in the data or whether the objects were only partitioned into some artificial groups \cite{Silhouettes}. Therefore, Silhouette Coefficient is a metric used to calculate the goodness of a clustering technique.

The silhouettes are helpful when the proximities are on a ratio scale and when seeking compact and separated clusters \cite{Silhouettes}. To construct silhouettes, it only needs two things: the partition by the application of some clustering technique and the collection of all proximities between objects. 

In Silhouette Coefficient, its value ranges from $-1$ to $1$. For values closer to $1$, the clusters are well apart and distinguished. On the other hand, they are indifferent to values more relative to $ 0 $, or the distance between them is insignificant. Finally, the clusters are assigned incorrectly for values closer to $-1$.

Equation (2) presents the formula for the Silhouette Coefficient. The $a$ is the average intra-cluster distance, i.e., the average distance between each point within a cluster, Equation (3), and $b$ is the average inter-cluster distance, i.e., the average distance between all partitions, Equation (4).

\begin{equation}
\frac{(b_{i}-a_{i})}{max(a_{i},b_{i})}
\end{equation}

\begin{equation}
a_{i} = \frac{1}{|C_{i}| - 1} \sum_{j \in C_{i}, i \neq j}d(i, j)
\end{equation}

\begin{equation}
b_{i} = \min_{k\neq i}\frac{1}{|C_{k}|} \sum_{j \in C_{k}}d(i, j)
\end{equation}

\subsection{Network Structure Inference}
Throughout history, the exploration of networks has predominantly belonged to the domain of discrete mathematics, more specifically, graph theory. This field began in 1735, with Euler's resolution of the problem of Königsberg bridges, resulting in the formulation of the concept of Eulerian graph \cite{Balaji}.

As defined by \cite{survey}, a network $G = (V, E, A)$ is a set composed of a set $V$ of $n$ nodes, where each node is represented by $v_{i} \in V$, a set $E$ of $m$ pairs of nodes, represented by $e_{ij} \in E$, and a set $A$ that contains information about nodes and/or sets of edge attributes. Among these attributes, the specific weight of an edge stands out, represented by $w_{ij}$, a scalar value that obeys the constraint $\vert w_{ij}\vert \leq 1$, $\vert w_{ij }\vert = 0$ indicating the absence of an edge. The network is classified as unweighted when $w_{ij} \in {0, 1}$. This last configuration represents a particular case of the weighted network. Furthermore, node and edge attributes assume a significant role as specific types of features derived from functions that measure local properties of the edge or node, such as, for example, the degree of the node. In turn, time-varying networks are conceptualized as sequences that contain snapshots of static networks: $G = (G_{1},...G_{k} ,...G_{t})$. These dynamic networks include attributes and/or edges that transform over time, creating a multifaceted panorama of the underlying interactions \cite{survey}.

The network topology inference problem merges at the intersection of two fundamental models. First, a network model, denoted as $R(D, \alpha) \to G$, builds the network representation $G$ from input data $D$ under a variety of parameters $\alpha$. These models can be exemplified by parametric statistical approaches, such as exponential random graph models, or by non-parametric and bounded interactions networks. Then, the problem incorporates a task model, expressed by $T (G, \beta) \to p_{1}, p_{2}...$, which acts on the network $G$ as input under specific parameters $\beta$, generating task results (such as "$p_{i}$" predictions). These results approximate the optimal hypothetical function $T^{*}(G)$ of a network task, based, for example, on classification or prediction, considering an error $e()$ \cite{survey}.

In this structure, it is feasible to carry out various tasks. In the context of network prediction tasks, the predictive model demonstrates abilities to generate predictions for (1) edges, (2) attributes, or (3) the data itself \cite{survey}. From a set of validation edges $E^{*}$, an application instance can be evaluated as in Equation $5$:

\begin{equation}
    \arg \min_{G} e(T(R(D, \alpha), \beta), E^{*})
\end{equation}

This implies the inference of $G$ based on network model parameters $\alpha$ and task model parameters $\beta$. In this scenario, the adequacy of the network and task models and the selection of a suitable error function is critical to determine the inferred network performance \cite{survey}. Network model selection methodologies mitigate some of the biases arising from the offline construction of manually tuned networks by fully exploring various combinations of possible models \cite{survey}.

This formulation highlights two fundamental challenges in the task of network structure inference. First, the parameter space of $G$, from $R(D, \alpha)$, can become substantially vast, and the exploration of these parameters, very possibly, will not present convexity about the performance of the task of interest. Second, models that perform well in this parameter space can result in remarkably distinct network topologies \cite{survey}. According to \cite{survey} synthesizing and harmonizing these discrepancies can be crucial to understanding the most appropriate network model. The wide variety of plausible tasks and network models generates a scenario that significantly challenges hypothesis generation and evaluation for the researcher; the presence of more credible models requires additional interpretive approaches to understanding the mechanisms underlying the system's behavior \cite{survey}.

\subsection{Graph Entropy}
The first researchers to define and explore the concept of graph entropy include \cite{77}, \cite{86}, \cite{66}. After these pioneering contributions, \cite{55} presented a distinct definition of graph entropy, strongly linked to issues in information theory and coding. This work aimed to solve the problem of evaluating the effectiveness of the ideal encoding of messages coming from an information source, in which the symbols belong to a finite set of vertices $V$. Another definition of Korner entropy, which first appeared in \cite{24}, is based on the well-known stable set problem and is closely related to minimum entropy colorings of graphs \cite{80}.

Network Entropy is based on the classic "Shannon Entropy" for discrete distributions \cite{Wiedermann}. \cite{Small} proposed a Network Entropy measure based on the probability of a random walker going from node $i$ to any other node $j$. According to \cite{FreitasCristopherGS2019Adco}, this probability distribution $P(i)$ is defined for each node $i$ as shown in Equation $6$. So $\sum_{j}p_{i \rightarrow j} = 1$

\begin{equation}
    p_{i \rightarrow j} = \left\{
    \begin{array}{lr}
        0, & for\ a_{ij} = 0\\
        \frac{1}{k_{i}}, & for\ a_{ij} = 1
    \end{array}
    \right.
\end{equation}

Based on the probability distribution $P^{(i)}$, the entropy for each node, with $\varphi^{(i)} = 0$, can be defined as in Equation $7$:

\begin{equation}
    \varphi^{(i)} \equiv  \varphi [P^{(i)}] = - \sum_{j=1}^{N- 1} p_{i \rightarrow j} \ln p_{i \rightarrow j} = \ln k_{i}
\end{equation}

Next, the normalized entropy of the node is calculated, Equation $8$:

\begin{equation}
     H^{(i)} = \frac{\varphi [P^{(i)}]}{\ln(N- 1)} = \frac{\ln k_{i}}{\ln (N- 1)}
\end{equation}

Finally, the normalized network entropy is calculated by averaging the normalized node entropy over the entire network, Equation $9$ \cite{FreitasCristopherGS2019Adco}:

\begin{equation}
    H = \frac{1}{N}\sum_{i = 1}^{N} H^{(i)} = \frac{1}{N \ln (N- 1)}  \sum_{i = 1}^{N} \ln k_{i}
\end{equation}

\section{Experimental Planning and Approaches}

\subsection{Scenarios}

The simulations were performed on the NetLogo \cite{Uri} platform because this platform has a library with simulation examples. In this library, we selected four simulations from the animal kingdom that had the characteristic of the organization. It was necessary to make some modifications to the simulations. Scenarios and changes are exemplified in the following subsections.

%We selected these four organizational structures, as only these structures were identified in the simulations of the animal kingdom made available by the Netlogo Platform.

\subsubsection{Ants}

\begin{figure}
\centering
\includegraphics[height=6cm]{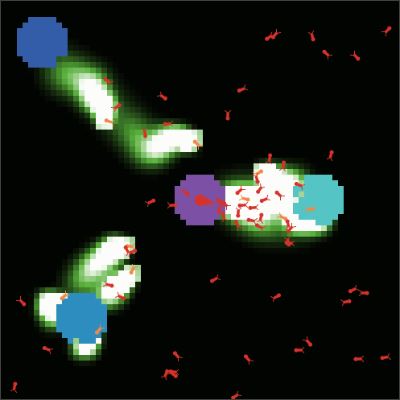}
\caption{Ants Simulation - Teams} \label{figure:ants}
\end{figure}

In this simulation, there is a colony of ants whose functions of each ant are to look for food in the environment and return the food to the anthill. When an ant finds a piece of food, it carries it back to the nest, dropping a chemical as it moves, Figure \ref{figure:ants}. When other ants "sniff" the chemical, they follow the chemical toward the food. As more ants carry food to the nest, they reinforce the chemical trail. 

The ant colony generally exploits the food source in order, starting with the food closest to the nest and finishing with the food most distant from the nest. It is more difficult for the ants to form a stable trail to the more distant food since the chemical trail has more time to evaporate and diffuse before being reinforced. Once the colony finishes collecting the closest food, the chemical trail to that food naturally disappears, freeing up ants to help collect the other food sources. However, the more distant food sources require a more significant "critical number" of ants to form a stable trail.

When the ants "take" all the food available in the scenario to the anthill, the organization is undone, and they walk randomly through the simulation scenario.

No one or more agents command the others. Instead, all the ants work together to reach the objective. Thus, this simulation contains the organizational paradigm of teams. To evaluate the organization detection in this scenario, we modified the ants' behavior so that they worked individually. That is, the ants look for food and take it to the anthill, but without informing the others. Thus, we can evaluate our approaches with and without organization.

\subsubsection{Wolf Sheep Predation}

\begin{figure}
\centering
\includegraphics[height=6cm]{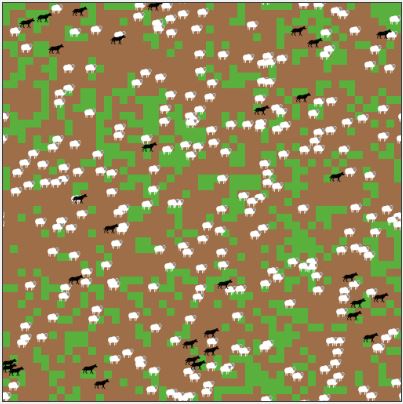}
\caption{Wolf Sheep Predation Simulation - Hierarchy} \label{figure:wolf}
\end{figure}

This model has two groups of agents: the wolves (black agents) and the sheep (white agents), Figure \ref{figure:wolf}. This simulation aims to explore the stability of predator-prey ecosystems. In this model, wolves and sheep wander randomly around the landscape while the wolves look for sheep to prey on. Each step costs the wolves and sheep energy, and they must eat to replenish their energy. If their run out of energy, they die. To allow the population to continue, each wolf or sheep has a fixed probability of reproducing at each time step. 

However, in this work, we wanted to evaluate our approach in several organizational paradigms. By knowing how wolves organize themselves to hunt their prey, in which an alpha wolf coordinates the attack on a target, we selected this simulation to simulate the hierarchical organization. Therefore, some modifications were necessary. In the original scenario, sheep and wolves die if their energy runs out and the wolves work individually. In our changes, only sheep die if a wolf eats them. As our approach involves observing the trajectories of agents that may contain some organization, in this case, the wolves do not die during the simulation. When the wolves "devour" all the sheep in the scenario, the group is disbanded and the wolves walk randomly.

In the hierarchical scenario, we added an alpha wolf agent that selects the prey for the other wolves to help him capture it. Whereas in the original scenario, all wolves work individually. Thus, we designed a predation scenario with and without wolf organization.

\subsubsection{Flocking}

\begin{figure}
\centering
\includegraphics[height=6cm]{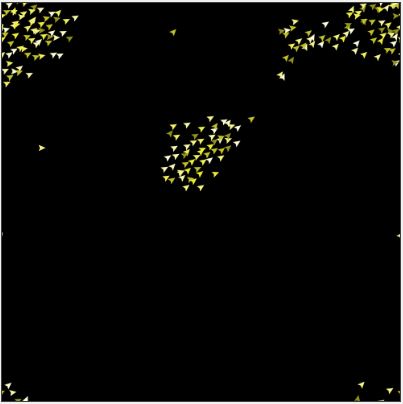}
\caption{Flocking Simulation - Coalition} \label{figure:flocking}
\end{figure}

This NetLogo template seeks to mimic flocking birds, Figure \ref{figure:flocking}. The resulting movement can also resemble schools of fish. Flocks spawned in this scenario are not created or led by any leader birds. Instead, each bird follows the same rules from which flocks arise and break up. Each flock is dynamic. Once together, a community is not guaranteed to keep all its members. 

The birds follow three rules: "alignment", "separation", and "cohesion". "Alignment" means that a bird tends to turn to move in the same direction that nearby birds are moving. "Separation" means that a bird will turn to avoid another bird that gets too close. "Cohesion" means that a bird will move towards other nearby birds (unless another bird is too close). When two birds are too close, the "separation" rule overrides the other two, deactivating until the minimum separation is achieved. The three rules affect only the bird's heading. Therefore, each bird always moves forward at the same constant speed.

As bands arise due to a need and disband when the requirement is fulfilled, the paradigm present in this simulation is the coalition. So, to generate a simulation without this organization to evaluate our approach, we modified the birds to roam randomly around the environment, with no intention of forming flocks to achieve their goals.

\subsubsection{Ants Adaptantion}

\begin{figure}
\centering
\includegraphics[height=3cm]{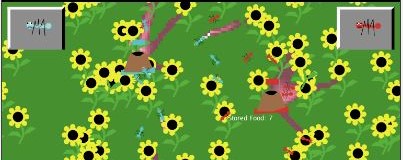}
\caption{Ants Adaptation Simulation - Congregation} \label{figure:ants2}
\end{figure}

In this simulation, there are two ant colonies (one of the red ants and one of the black ants) and many flowers are created in the world, Figure \ref{figure:ants2}. Ants are spawned in the ant colony and then wander until they find food, represented as flowers in the model. When they find food, they gain energy through eating nectar. Then they return to the colony while laying down a pheromone. Ants near pheromone trails are attracted to the most potent chemical trail. As the ants exhaust a food source, they once again begin to wander until they find another food source or pheromone trail to follow.

When two or more ants of opposing colonies encounter each other, they fight or scare each other away, leaving chemicals that attract more ants. For the winner, this works to protect the food source from competing colonies. The ant queen reproduces when the ants in her colony collect enough food, set by the created cost for each colony. Flowers periodically grow up around the map, resupplying food in the game. Ants die if they get too old, cannot find food, or sometimes when they lose a fight. Finally, nests die if they have no more ants living in them.

Unlike Ants simulation, ants can die or spawn new ants. Therefore, just like the Wolf Sheep Predation simulation, as our goal is to follow the agents' trajectories to identify patterns of organization, we modified this simulation so that the agents in this scenario are fixed, i.e., they do not die, and ones are not generated. Like the Ants simulation, when there is no more food in the simulation scenario, the organization is undone, and they walk randomly in the environment.

As the rest remains as in the original scenario, the paradigm of this organization is the congregation since the colonies are long-term and not created by a simple objective that when supplied, the colonies dissolve. Furthermore, there is not one agent leading the others in each colony. So, to remove this paradigm from this simulation, we performed the same modification as in the Ants scenario. Ants will look for food and return it to the anthill, but they do not collaborate on the food location; they work individually.

\subsection{Proposed Approaches}

\subsubsection{Adjacency Matrices}

Initially, our approach obtains the trajectory matrix, $M_{traj}$, of all agents during the simulation period, $time$. The initial design of the algorithm was based on graphs, whereby collecting $X$ simulation steps, a trajectory of $X$ steps is obtained for each agent. A graph is constructed where each path is a vertex. For $Y$ agents, we will have a graph with $Y$ vertices. Next, the similarity of each vertex is evaluated.

The similarity was evaluated by trajectory similarity and distance between vertices. We use cosine similarity to determine how similar the trajectories are. The distance was used, as the vertices may be identical but are far from each other, so there was no interaction between the agents even though they presented a similar path. The distances between each vertex were normalized in a scale from $0$ to $1$. 

Obtaining the $Ms$ matrix, which are adjacency matrices, with the similarity values of the trajectories at index 0 ($Ms[0]$) and the distance of each agent during the simulation at index 1 ($Ms[1]$), we apply the metric described in Equation $10$, than these two values, to verify the degree of similarity between the agents. This generates a final similarity matrix $num\_agents\times num\_agents$ ($M_{yes}$), where $num\_agents$ is the number of agents in the simulation.

\begin{equation}
M_{sim} = (1 - \mathrm{mod}(Ms[0])) * Ms[1]
\end{equation}

\subsubsection{Cluster Quality Indexes}

The similarity matrix is used in training the DBSCAN algorithm to generate the clusters. Finally, the Silhouette Coefficient is applied to assess the quality of the partitions in scenarios where there are and where there are no organizations.

Summarizing the description made in the previous paragraphs about our algorithm, Algorithm 2 shows the pseudocode. The input parameters are the trajectory matrix, the simulation time, the window size, and the number of agents in the simulation. For each time step, the trajectory similarity of the agents is calculated for a period (window size) based on the similarity of the trajectory and distance between the agents. After that, clusters of similar agents are generated in the given period of time, and the quality of the clusters is evaluated.

\begin{algorithm}[H]
  \caption{Detecting Organization by Cluster Quality Indexes}
  \scriptsize
  \label{algorithmcluster}
  \begin{algorithmic}[1]
  \Require $M_{traj}, time, size\_window, num\_agents$
    \Function{CalculeSimilatity}{$start$, $size\_window$, $M_{traj}$, $num\_agents$}
        \For{$k\in\{1,...,num\_agents\}$}
            \State $windowing[0] = M_{traj}[start,start + size\_window, k, 0]$
            \State $windowing[1] = M_{traj}[start,start + size\_window, k, 1]$
            \State $M_{window} = windowing$
        \EndFor
        
        \State $Ms[0] \gets cosine\_similarity(M_{window})$ 
        \State $Ms[1] \gets pairwise\_distances(M_{window})$
        \State $Ms[1] \gets normalize(Ms[1])$
      
      \Return $Ms$
    \EndFunction
    
    \For{$k\in\{1,..., time - size\_window\}$}
        \State $Ms \gets CalculeSimilatity(k, size\_window, M_{traj}, num\_agents)$
        \State $M_{sim} \gets (1 - \mathrm{mod}(Ms[0])) \circ Ms[1]$
        \State $clusters \gets DBSCAN(M_{sim})$
        \State $IndClusters[k] \gets silhouette\_score(clusters)$
    \EndFor
  \end{algorithmic}
\end{algorithm}

\subsubsection{Graph Entropy}

In the graph entropy approach, after obtaining the adjacency matrix, we define the number of nodes that equals the number of agents in each simulation and calculate the normalized network entropy based on the formula of Equations $7, 8 and 9$, as shown in Algorithm $3$.

Algorithm 2 shows the pseudocode this approach. As well as the approach with the cluster quality index, the input parameters are the trajectory matrix, the simulation time, the window size, and the number of agents in the simulation. For each time step, the trajectory similarity of the agents is calculated for a period (window size) based on the similarity of the trajectory and distance between the agents. After that, the entropy of the graph generated by the adjacency matrix is calculated in the time period of the window size.

\begin{algorithm}[H]
  \caption{Detecting Organization by Graph Entropy}
  \scriptsize
  \label{algorithmentropy}
  \begin{algorithmic}[1]
  \Require $M_{traj}, time, size\_window, num\_agents$
    \Function{CalculeSimilatity}{$start$, $size\_window$, $M_{traj}$, $num\_agents$}
        \For{$k\in\{1,...,num\_agents\}$}
            \State $windowing[0] = M_{traj}[start,start + size\_window, k, 0]$
            \State $windowing[1] = M_{traj}[start,start + size\_window, k, 1]$
            \State $M_{window} = windowing$
        \EndFor
        
        \State $Ms[0] \gets cosine\_similarity(M_{window})$ 
        \State $Ms[1] \gets pairwise\_distances(M_{window})$
        \State $Ms[1] \gets normalize(Ms[1])$
      
      \Return $Ms$
    \EndFunction
    
    \For{$k\in\{1,..., time - size\_window\}$}
        \State $Ms \gets CalculeSimilatity(k, size\_window, M_{traj}, num\_agents)$
        \State $num\_nodes \gets num\_agents$
        \State $normalized entropy \gets \frac{1}{num\_nodes\times\ln(num\_nodes - 1)} \times \ln(\sum Ms)$
    \EndFor
  \end{algorithmic}
\end{algorithm}

\subsubsection{Experiment Parameters}

In the attempt to evaluate the approach in several paradigms with different amounts of agents, in scenarios with few agents, there was no cluster generated, i.e., all agents were labeled as noise in the simulation, which caused an error in the Silhouette Coefficient. To work around this problem, missing values (NaN) were added to the quality score vector when this problem occurred. Table 1 presents all configurations used in the experiments of this work. The dimension of the scenario for each simulation was the default configured by the Netlogo library.

\begin{table}
\centering
\caption{Experiment Setup.}\label{tab1}
\scriptsize
\begin{tabular}{|l|c|}
\hline
Parameters &  Values\\
\hline
Ants (Scenarios)&  150 ants\\
Wolf (Scenarios) &  15 wolves\\
Flocking (Scenarios) & 300 birds\\
Ants Adaptation (Scenarios) & 10 blue ants and 10 red ants\\
Simulation Time (Scenarios) & 500 time steps\\
Eps (DBSCAN) & $0.01$\\
Min\_samples (DBSCAN) & 5 \\
Metric (DBSCAN) & precomputed\\
Metric (Silhouette Score) & euclidean\\
$size\_{window}$ & 25 and 50 \\
\hline
\end{tabular}
%\vspace{-14mm}
\end{table}

The value of $500$ time steps was selected because, in simulations with a goal, such as those of ants, wolves, and sheep, this was the average time for agents in these simulations to reach their goals. Afterward, the agents randomly walk around the environment and match the simulations without organization. In addition, window sizes of $25$ and $50$ were chosen due to the values defined in the study conducted by \cite{related1}.

\section{Results and Analysis}

This section presents the results obtained with our approaches in the four organization simulation scenarios available in the NetLogo platform. We use a smoothed curve in the graphs to visualize better and interpret the results. Based on the configurations of the simulations used in this paper to evaluate the approaches, that is, number of agents, size of agents, world with no limit or with limit, and size of the simulation scenario, and on the results observed in the simulations.

\subsection{Cluster Quality Indexes}

\begin{figure}[ht]
\centering
\begin{subfigure}{.48\textwidth}
  \centering
  \includegraphics[width=\linewidth]{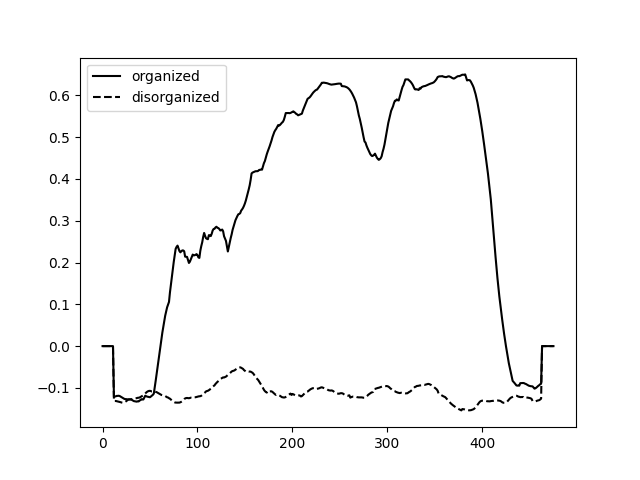}  
  \caption{Window size = 25}
  \label{fig:1}
\end{subfigure}\hfill
\begin{subfigure}{.48\textwidth}
  \centering
  \includegraphics[width=\linewidth]{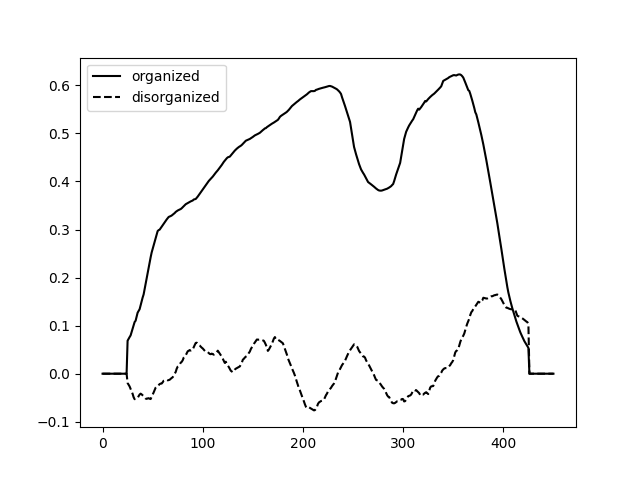}  
  \caption{Window size = 50}
  \label{fig:2}
\end{subfigure}
\caption{Quality index of Ants simulation clusters with and without organization. The y-axis corresponds to the cluster quality index and the x-axis corresponds to the simulation time period.}
\label{results:ants}
\end{figure}

\begin{figure}[ht]
\begin{subfigure}{.48\textwidth}
  \centering
  % include first image
  \includegraphics[width=1\linewidth]{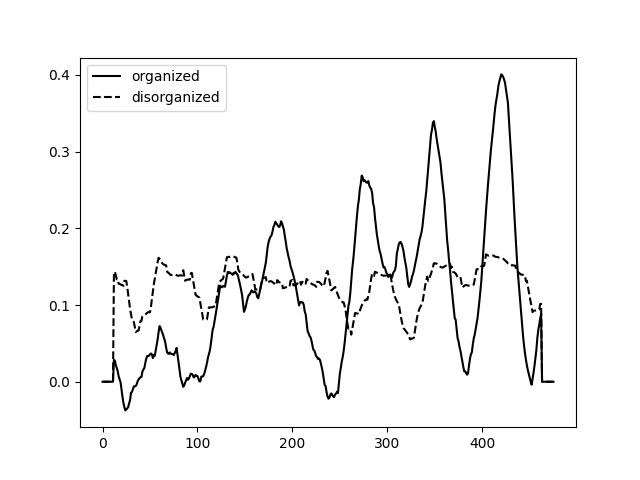}  
  \caption{window size = 25}
  \label{fig:3}
\end{subfigure}
\begin{subfigure}{.48\textwidth}
  \centering
  % include second image
  \includegraphics[width=1\linewidth]{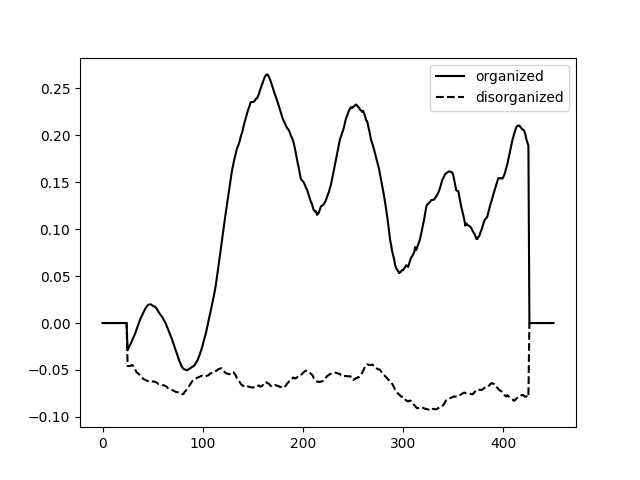}  
  \caption{window size = 50}
  \label{fig:4}
\end{subfigure}
\caption{Quality index of Flocking simulation clusters with and without organization. The y-axis corresponds to the cluster quality index and the x-axis corresponds to the simulation time period.}
\label{results:flocking}
\end{figure}

\begin{figure}[ht]
\begin{subfigure}{.48\textwidth}
  \centering
  % include first image
  \includegraphics[width=1\linewidth]{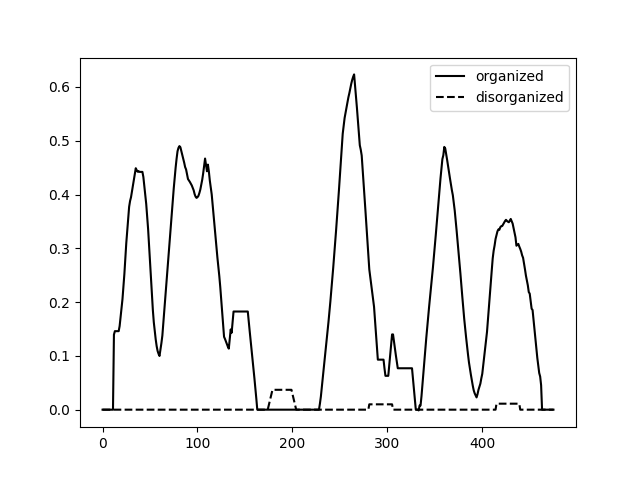}  
  \caption{window size = 25}
  \label{fig:5}
\end{subfigure}
\begin{subfigure}{.48\textwidth}
  \centering
  % include second image
  \includegraphics[width=1\linewidth]{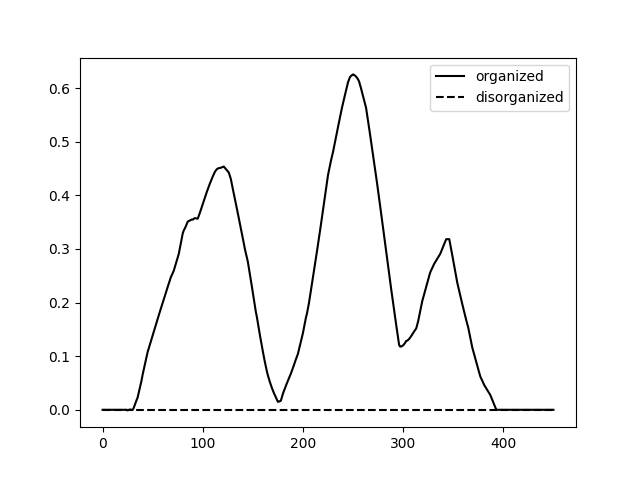}  
  \caption{window size = 50}
  \label{fig:6}
\end{subfigure}
\caption{Quality index of Wolf Sheep Predation simulation clusters with and without organization. The y-axis corresponds to the cluster quality index and the x-axis corresponds to the simulation time period.}
\label{results:wolf}
\end{figure}

\begin{figure}[ht]
\begin{subfigure}{.48\textwidth}
  \centering
  % include first image
  \includegraphics[width=1\linewidth]{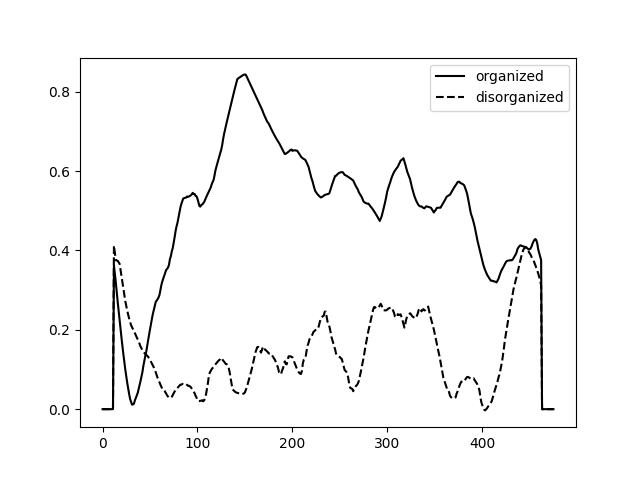}  
  \caption{window size = 25}
  \label{fig:7}
\end{subfigure}
\begin{subfigure}{.48\textwidth}
  \centering
  % include second image
  \includegraphics[width=1\linewidth]{Figuras/ants_50.png}  
  \caption{window size = 50}
  \label{fig:8}
\end{subfigure}
\caption{Quality index of Ants Adaptation simulation clusters with and without organization. The y-axis corresponds to the cluster quality index and the x-axis corresponds to the simulation time period.}
\label{results:antsadap}
\end{figure}

Figures \ref{results:ants}, \ref{results:flocking}, \ref{results:wolf}, and \ref{results:antsadap} show the results of the quality indices of the clusters generated during the $500$ time steps, with window size equal to $25$ and $50$, in the Ants (Teams), Flocking (Coalitions), Wolf Sheep Predation (Hierarchy) and Ants Adaptation (Congregations), respectively. The Figures contain the results of the simulations with and without organizations.

In Figure \ref{results:ants} it is possible to notice that the quality indices in most simulation times were higher than $0$ in the scenarios with organized ants. In this scenario, values close to $0$ occur when no food is detected and ants walk randomly to locate it. When the pheromone is found, it is released and the ants begin to cluster in the indicated food path. Thus, the value of the clusters' quality index increases. On the other hand, in scenarios with disorganized ants, the quality indices, in most simulation times, were lower or close to $0$ because there is no cooperation/interaction between the ants, they pick up the food they randomly find and take it to the anthill, without releasing pheromones, so there is no grouping in the path of food.

The Figure \ref{results:flocking} presents the results of the quality scores of the Flocking simulation. In this scenario, there are many agents, just like in the Ants simulation, but the environment is limitless. The agents in a simulation can move freely in any direction without being impeded by edges or borders. They reappear in the opposite direction as if they had continued the movement. Due to this characteristic, the quality index of clusters with organized birds was lower than that of others simulations. However, when compared with the simulations with disorganized agents, it is possible to observe in the window with size $50$ that the quality index in the organized simulation has higher values, while in the case of disorganized agents the index was lower than $0$ in most of the simulation time.

%\begin{figure}
%\centering
%\includegraphics[height=7cm]{Figuras/flocking-coalizao.png}
%\caption{Quality index of Flocking simulation clusters with and without organization. The y-axis corresponds to the cluster quality index and the x-axis corresponds to the simulation time period.} \label{results:flocking}
%\end{figure}

In the simulation with the wolves, Figure \ref{results:wolf}, these are low-agent scenarios compared to Ants and Flocking simulations. Thus, clusters may not be formed because all the agents were classified as noise for the low interaction, especially in environments without organization. However, with organized wolves, it was still possible to obtain better results than in Flocking, with the quality index reaching $0.6$ in both window sizes, as the world of this simulation is limited, and the wolves do not separate at any time. Instead, they always walk in a pack after the next prey. However, when the wolves were disorganized, there were few interactions between them, so the cluster quality index in most simulations was equal to $0$ most of the time.

Finally, in Figure \ref{results:antsadap}, it is possible to notice that the quality of the clusters was superior to $0$ for the organized ants, obtaining a quality superior to $0.6$ in some moments, even having few agents in the simulations. While in disorganized ants, the value was equal to or close to $0$. In just one instant the value was less than $0$. This occurs because the environment has a smaller dimension and the size of the agents are larger when compared to the other scenarios, allowing greater interaction, especially in organized scenarios.

\subsection{Graph Entropy}

\begin{figure}[ht]
\begin{subfigure}{.48\textwidth}
  \centering
  % include first image
  \includegraphics[width=1\linewidth]{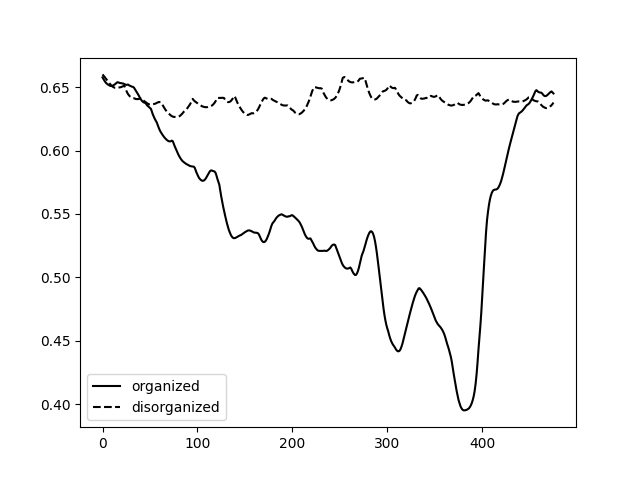}  
  \caption{window size = 25}
  \label{fig:9}
\end{subfigure}
\begin{subfigure}{.48\textwidth}
  \centering
  % include second image
  \includegraphics[width=1\linewidth]{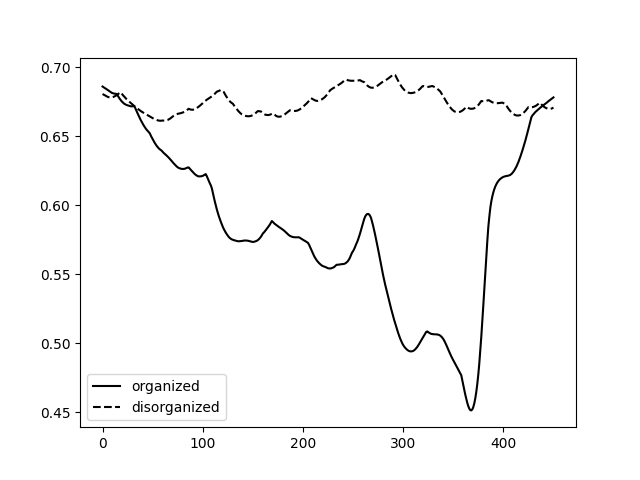}  
  \caption{window size = 50}
  \label{fig:10}
\end{subfigure}
\caption{Graphs entropy of Ant simulation with and without organization. The y axis corresponds to the entropy value and the x axis corresponds to the simulation period.}
\label{results:ants_entropy}
\end{figure}

\begin{figure}[ht]
\begin{subfigure}{.48\textwidth}
  \centering
  % include first image
  \includegraphics[width=1\linewidth]{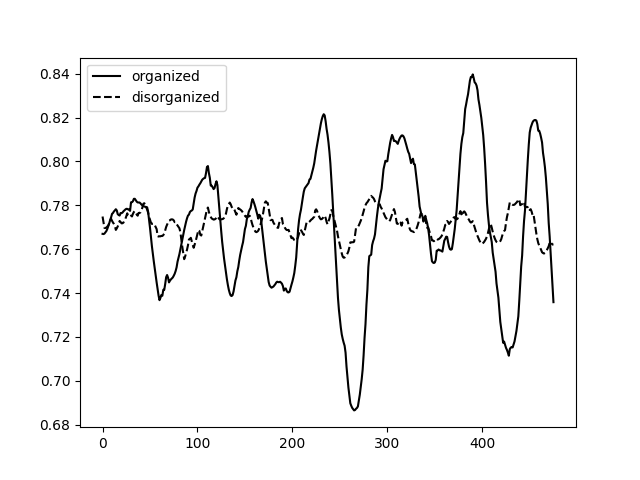}  
  \caption{window size = 25}
  \label{fig:11}
\end{subfigure}
\begin{subfigure}{.48\textwidth}
  \centering
  % include second image
  \includegraphics[width=1\linewidth]{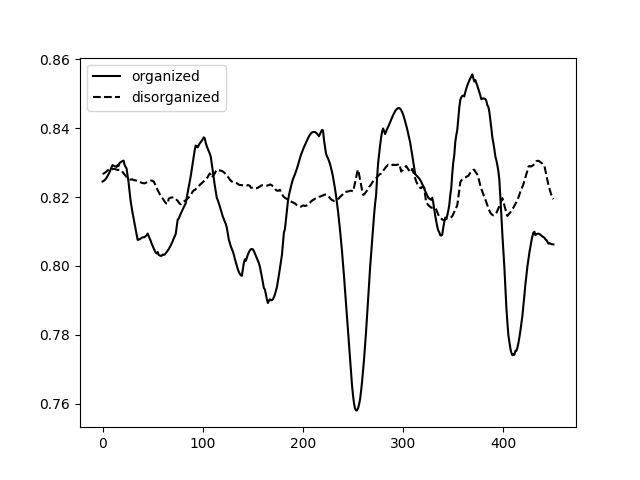}  
  \caption{window size = 50}
  \label{fig:12}
\end{subfigure}
\caption{Graphs entropy of Flocking simulation with and without organization. The y axis corresponds to the entropy value and the x axis corresponds to the simulation period.}
\label{results:flocking_entropy}
\end{figure}

\begin{figure}[ht]
\begin{subfigure}{.48\textwidth}
  \centering
  % include first image
  \includegraphics[width=1\linewidth]{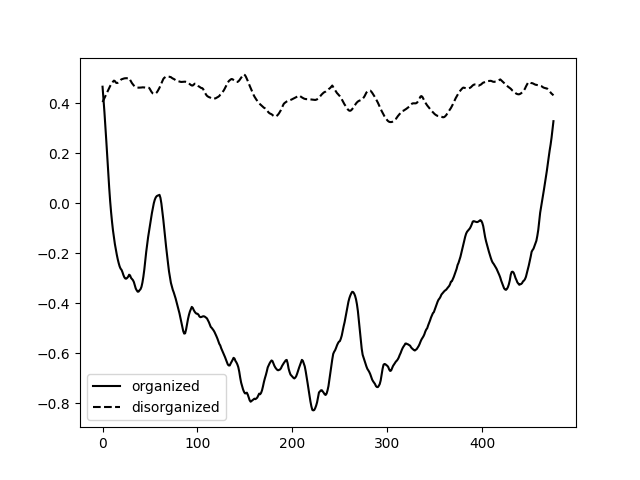}  
  \caption{window size = 25}
  \label{fig:13}
\end{subfigure}
\begin{subfigure}{.48\textwidth}
  \centering
  % include second image
  \includegraphics[width=1\linewidth]{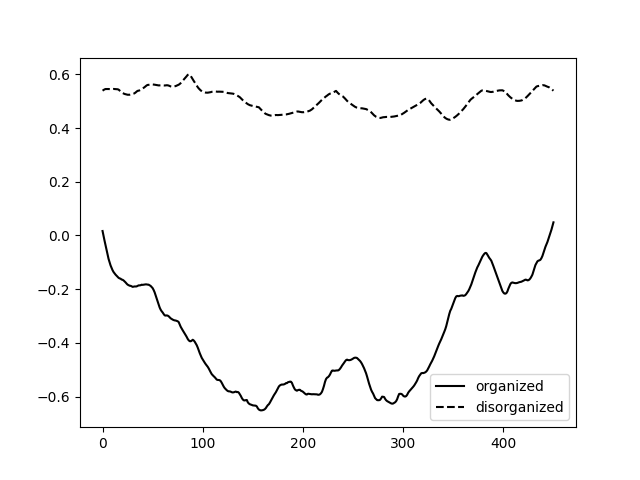}  
  \caption{window size = 50}
  \label{fig:14}
\end{subfigure}
\caption{Graphs entropy of Wolf Sheep Predation simulation with and without organization. The y axis corresponds to the entropy value and the x axis corresponds to the simulation period.}
\label{results:wolf_entropy}
\end{figure}

\begin{figure}[ht]
\begin{subfigure}{.48\textwidth}
  \centering
  % include first image
  \includegraphics[width=1\linewidth]{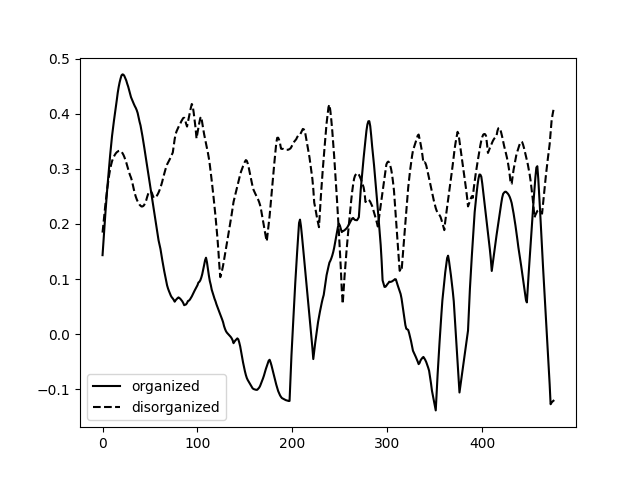}  
  \caption{window size = 25}
  \label{fig:15}
\end{subfigure}
\begin{subfigure}{.48\textwidth}
  \centering
  % include second image
  \includegraphics[width=1\linewidth]{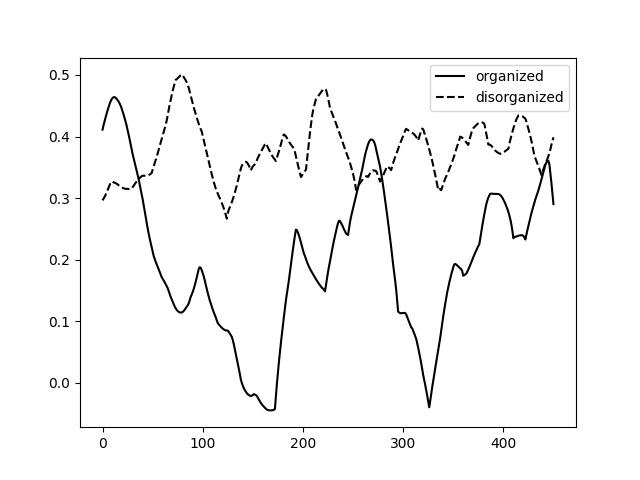}  
  \caption{window size = 50}
  \label{fig:16}
\end{subfigure}
\caption{Graphs entropy of Ants Adaptation simulation with and without organization. The y axis corresponds to the entropy value and the x axis corresponds to the simulation period.}
\label{results:antsadap_entropy}
\end{figure}

Figures \ref{results:ants_entropy}, \ref{results:flocking_entropy}, \ref{results:wolf_entropy}, and \ref{results:antsadap_entropy} show the results of the graph entropy generated during the $500$ time steps, with window size equal to $25$ and $50$, in the Ants (Teams), Flocking (Coalitions), Wolf Sheep Predation (Hierarchy) and Ants Adaptation (Congregations), respectively. The Figures contain the results of the simulations with and without organizations. A graph with entropy close to $0$ suggests that the connections between nodes are highly deterministic, that is, there is a clear trend or pattern in the relationships \cite{related2}.

In Figure \ref{results:ants_entropy}, it is possible to notice that in the organized simulations the entropy values were lower when compared to the disorganized simulations. As well as in the approach with cluster quality indexes, in the simulations with organization the entropy values were higher at the beginning when no food was yet located and at the end when the food runs out. 

The Figure \ref{results:flocking_entropy} presents the results of the graph entropy of the Flocking simulation. Due the characteristic this simulation (there are many agent and the environment is limitless), entropy in the simulation with disorganization maintained constant entropy values, with more discreet peaks, while in the organized simulation there are more visible peaks where entropy is lower than the disorganized one, but there are also peaks of higher entropy in the two configured window sizes.

In the simulation with the wolves, Figure \ref{results:wolf_entropy}, as well as in the Ants simulation, it is possible to notice that in the scenario with organized agents the entropy values were lower than in the scenario with disorganized agents. This difference is more noticeable with the window equal to $50$.

In Figure \ref{results:antsadap}, even with close values, it is possible to notice that the entropy was lower for the organized agents, especially with the window equal to $50$.

\subsection{Wolfgang Trumler and Mike Gerdes' Approach}

Figures \ref{results:ants_entropy}, \ref{results:flocking_entropy}, \ref{results:wolf_entropy}, and \ref{results:antsadap_entropy} present the results of the simulations seeking to detect patterns that evidence organized agents or not using the approach proposed by \cite{related1} to use entropy calculation in x and y coordinates.

\begin{figure}[ht]
\begin{subfigure}{.48\textwidth}
  \centering
  % include first image
  \includegraphics[width=1\linewidth]{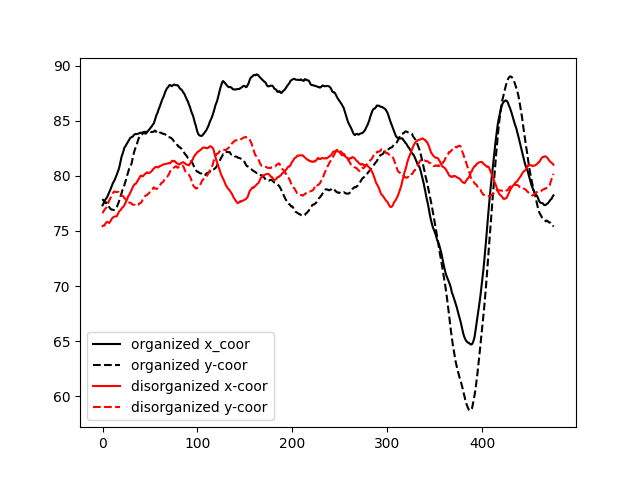}  
  \caption{window size = 25}
  \label{fig:17}
\end{subfigure}
\begin{subfigure}{.48\textwidth}
  \centering
  % include second image
  \includegraphics[width=1\linewidth]{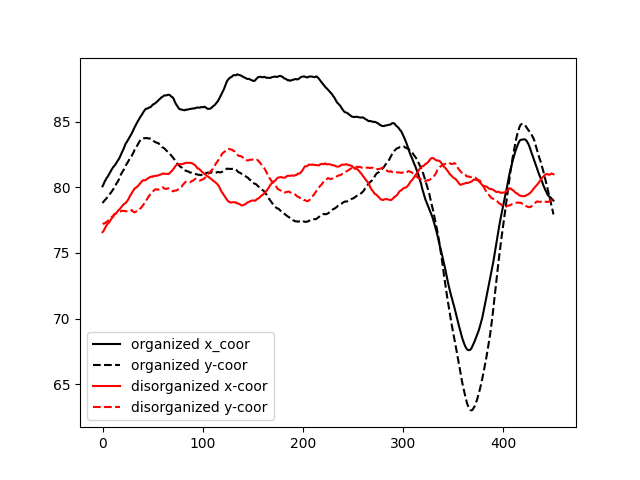}  
  \caption{window size = 50}
  \label{fig:18}
\end{subfigure}
\caption{Wolfgang Trumler and Mike Gerdes' approach in Ant simulation with and without organization. The y axis corresponds to the entropy value and the x axis corresponds to the simulation period.}
\label{results:ants_entropy_related}
\end{figure}

In Figure \ref{results:ants_entropy_related}, a peak occurred in which entropy was lower in the organized scenario than the disorganized one. However, in general, there is no clear vision of a pattern that allows the differentiation of each case.

\begin{figure}[ht]
\begin{subfigure}{.48\textwidth}
  \centering
  % include first image
  \includegraphics[width=1\linewidth]{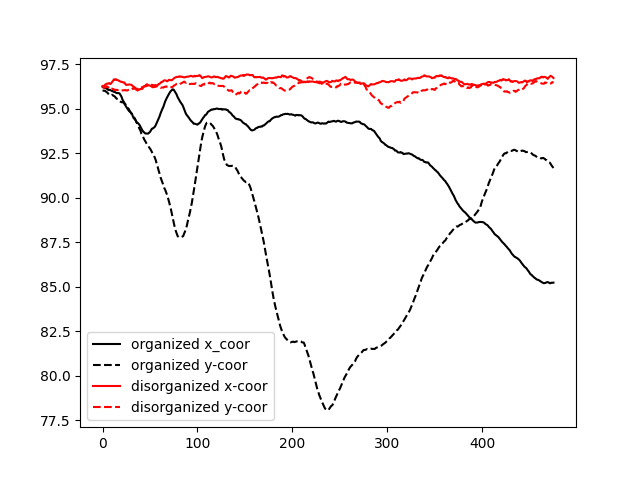}  
  \caption{window size = 25}
  \label{fig:19}
\end{subfigure}
\begin{subfigure}{.48\textwidth}
  \centering
  % include second image
  \includegraphics[width=1\linewidth]{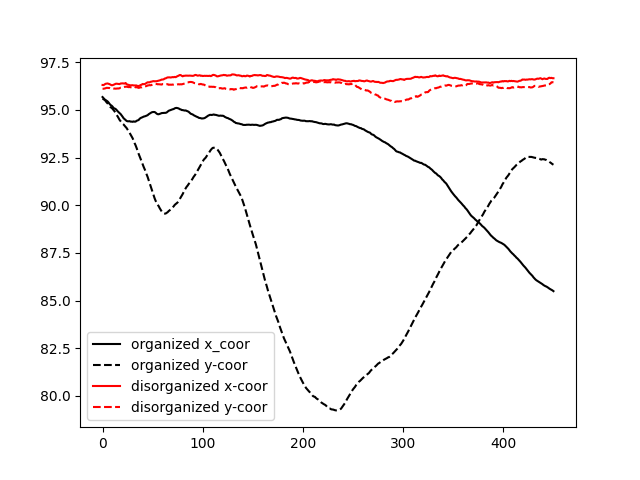}  
  \caption{window size = 50}
  \label{fig:20}
\end{subfigure}
\caption{Wolfgang Trumler and Mike Gerdes' approach in Flocking simulation with and without organization. The y axis corresponds to the entropy value and the x axis corresponds to the simulation period.}
\label{results:flocking_entropy_related}
\end{figure}

The Figure \ref{results:flocking_entropy_related} presents the results of the approach by \cite{related1} in the Flocking simulation. Different from the Ants simulation, in this simulation it is possible to identify the difference of each scenario. Because while the value of entropy is constant in the disorganized scenario, with organized agents there are peaks with lower values, especially in the y coordinate.

\begin{figure}[ht]
\begin{subfigure}{.48\textwidth}
  \centering
  % include first image
  \includegraphics[width=1\linewidth]{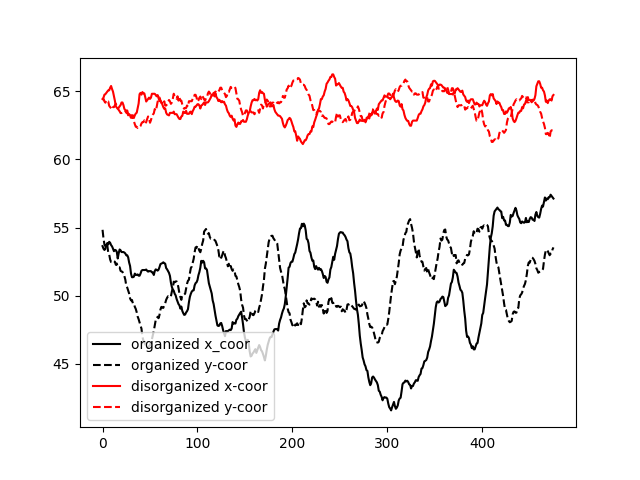}  
  \caption{window size = 25}
  \label{fig:21}
\end{subfigure}
\begin{subfigure}{.48\textwidth}
  \centering
  % include second image
  \includegraphics[width=1\linewidth]{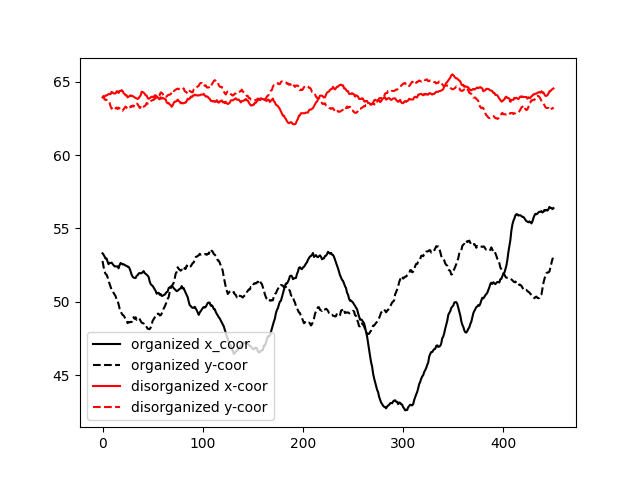}  
  \caption{window size = 50}
  \label{fig:22}
\end{subfigure}
\caption{Wolfgang Trumler and Mike Gerdes' approach in Wolf Sheep Predation simulation with and without organization. The y axis corresponds to the entropy value and the x axis corresponds to the simulation period.}
\label{results:wolf_entropy_related}
\end{figure}

In the simulation with the wolves, Figure \ref{results:wolf_entropy_related}, just like the Flocking simulation, the differentiation of the entropy of each scenario is more visible. However, unlike the Flocking simulation, it is possible to observe this in the two coordinates, x and y.

\begin{figure}[ht]
\begin{subfigure}{.48\textwidth}
  \centering
  % include first image
  \includegraphics[width=1\linewidth]{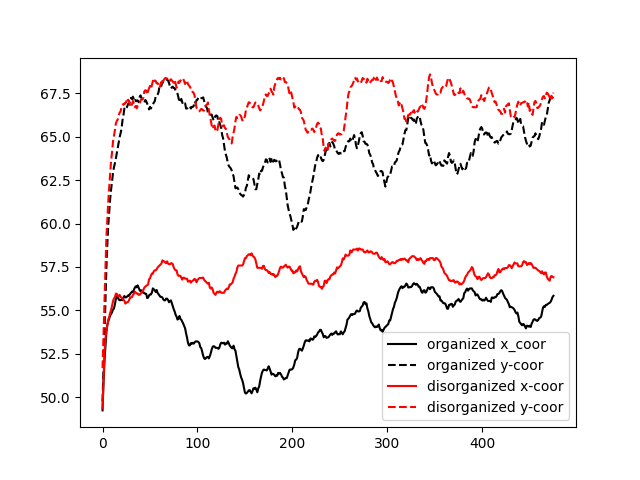}  
  \caption{window size = 25}
  \label{fig:23}
\end{subfigure}
\begin{subfigure}{.48\textwidth}
  \centering
  % include second image
  \includegraphics[width=1\linewidth]{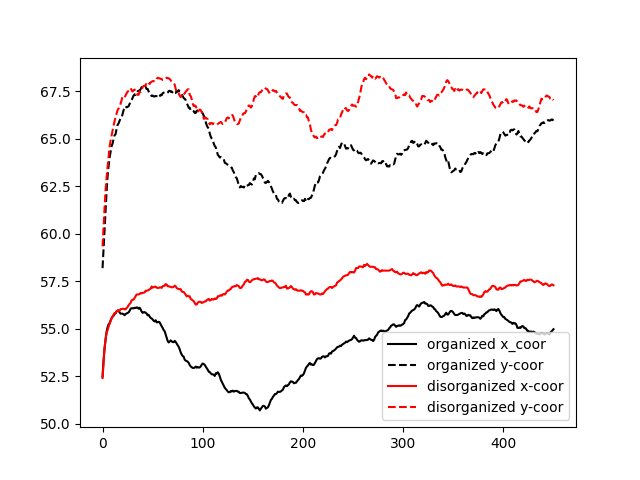}  
  \caption{window size = 50}
  \label{fig:24}
\end{subfigure}
\caption{Wolfgang Trumler and Mike Gerdes' approach in Ants Adaptation simulation with and without organization. The y axis corresponds to the entropy value and the x axis corresponds to the simulation period.}
\label{results:antsadap_entropy_related}
\end{figure}

In Figure \ref{results:antsadap_entropy_related}, even though it is possible to notice the difference in the graph, the entropy values of each coordinate for each scenario are close.

\subsection{Overall Result}

What can we see in the results of our approach using cluster quality indices, Figures \ref{results:ants}, \ref{results:flocking}, \ref{results:wolf}, and \ref{results:antsadap}, is that this approach made it possible to more clearly identify each scenario, the organized and the disorganized. Even if in some simulations the results are better, in all simulations it is observable that in disorganized scenarios the cluster quality indexes are below or close to $0$, while the organized scenarios obtained higher values of this index, considering the size of the window equal to $50$. It is worth mentioning that the number of agents in the simulation (scenarios as more agents tend to generate more clusters), the size of the simulation scenario (smaller scenarios tend to bring agents closer together), limited simulation environment (more open environments, in which agents can leave the simulation scenario, can lead to less cluster generation), etc. 

The results of our approach using graph entropy showed that in some simulations the patterns, Figures \ref{results:ants_entropy} and \ref{results:wolf_entropy} were more visible than in others that do not present a constant difference of values, but peaks where the values are lower than the scenarios with agents working individually, Figures \ref{results:flocking_entropy} and \ref{results:antsadap_entropy}. The same occurs with the approach of Wolfgang Trumler and Mike Gerdes, in the Flocking and Wolf Sheep Predation simulations it is more visible to identify each scenario, Figure \ref{results:flocking_entropy_related} and \ref{results:wolf_entropy_related}, while in the Ants and Ants simulations Adaptation there are either peaks or the values are very close, Figures \ref{results:ants_entropy_related} and \ref{results:antsadap_entropy_related}. However, our approach with graph entropy obtained values closer to 0 than the approach proposed by \cite{related1}.

\section{Conclusion}

This article proposes two approaches for detecting evidence of group organization. We use simulations of the animal world from the Netlogo platform to demonstrate the performance of our proposed approaches to demonstrate their applicability in the real world. The proposed approaches consist of using the index of clustering quality and entropy of the graphs in the adjacency matrix that contains information about the trajectory similarity and the distance between the agents at each simulation time step to identify organizational characteristics.

Compared with the approach proposed by \cite{related1} in the literature, the results showed that the approach with the cluster quality index allowed a better identification of each scenario (organized or disorganized) when compared to the other approaches presented in this article. However, our approach with graph entropy was better than the one proposed by \cite{related1}.

However, it is essential to note that the number of agents in the environment can affect the quality of the clusters, as all agents can be labeled as noise by DBSCAN. Thus, in future works, we intend to improve the algorithm, increasing more characteristics of the organization to be identified. Furthermore, we intend to evaluate our approach in real scenarios to validate its applicability and improve the detection of organizations. We also plan to apply the approach to scenarios with groups organized between individuals walking randomly, seeking to simulate real public scenarios to validate the applicability of our approach.

\section*{Acknowledgments}
This was was supported in part by......

%Bibliography
\bibliographystyle{unsrt}  
\bibliography{references}

\end{document}